%% file: Gasmarket.tex
\documentclass{llncs}
\usepackage{enumerate}
\usepackage{graphicx}
\usepackage{subfigure}
\usepackage{enumerate}
\usepackage{algorithmic, algorithm}
\usepackage{float}
\usepackage{array, listings}
\usepackage{amsmath}

\begin{document}

\title{Agent-based modeling of a price information trading business}

\author{Saad Ahmad Khan and Ladislau B{\"o}l{\"o}ni}
\institute{Department of Electrical Engineerng and Computer Sciences
      \\ University of Central Florida
      \\ 4000 Central Florida Blvd, Orlando FL 32816\\
       \email{\{skhan, lboloni\}@eecs.ucf.edu}}

\maketitle
\begin{abstract}

We describe an agent-based simulation of a fictional (but feasible)
information trading business. The Gas Price Information Trader (GPIT)
buys information about real-time gas prices in a metropolitan area from
drivers and resells the information to drivers who need to refuel their
vehicles.

Our simulation uses real world geographic data, lifestyle-dependent
driving patterns and vehicle models to create an agent-based model of
the drivers. We use real world statistics of gas price fluctuation to
create scenarios of temporal and spatial distribution of gas prices. The
price of the information is determined on a case-by-case basis through a
simple negotiation model. The trader and the customers are adapting
their negotiation strategies based on their historical profits.

We are interested in the general properties of the emerging information
market: the amount of realizable profit and its distribution between the
trader and customers, the business strategies necessary to keep the
market operational (such as promotional deals), the price elasticity of
demand and the impact of pricing strategies on the profit.

\end{abstract}

\input{introduction}

\input{profitsDiscussion}

\input{Negotiation}

\input{dataset}

\input{experiment}

\input{relatedWork}

\input{conclusion}

\bibliographystyle{abbrv}
\bibliography{refs}  

\end{document}

%% file: introduction.tex
\section{Introduction} 

In this paper we describe an agent-based simulation of a fictional (but
feasible) information trading business. The Gas Price Information Trader
(GPIT) buys information about real-time gas prices in a metropolitan
area from drivers, and resells this information to drivers who need to
refuel their vehicles. An autonomous agent, potentially integrated with
the vehicle's onboard control system, can in the near future perform all
the actions associated with the driver, including the acquisition of gas
prices, selling and buying information and negotiating with the trader.

Similar systems have been proposed previously
\cite{bulusu2008participatory,dong2008automatic}, online webpages
tracking gas prices currently exist (the ad-supported GasBuddy and
GasPriceWatch in the United States, the government-run FuelWatch in
Western Australia). These are related in spirit and operation to a large
number of applications proposed in the field of urban computing /
citizen computing, which can be similarly construed as information
trading. While feasibility has been repeatedly demonstrated, many of
these systems have not been, in general, deployment successes. We argue
that the low deployment is due to the fact that voluntary participation
can be only maintained for projects with emotional and political impact. Projects involving the environment, pollution, conservation can gather significant following. A gas price reporting system, however, does not have an emotional motivation for contribution, increasing the chance of free riding. 

Instead of relying on voluntary participation, our system relies on the self interest of the participants. We assume that the business is stable only if all the participants profit financially over the long run. We are interested in the overall market dynamics of the system.  

\begin{figure}[h!]
\vspace{1cm}
\centerline{
\includegraphics[width=3.5in]{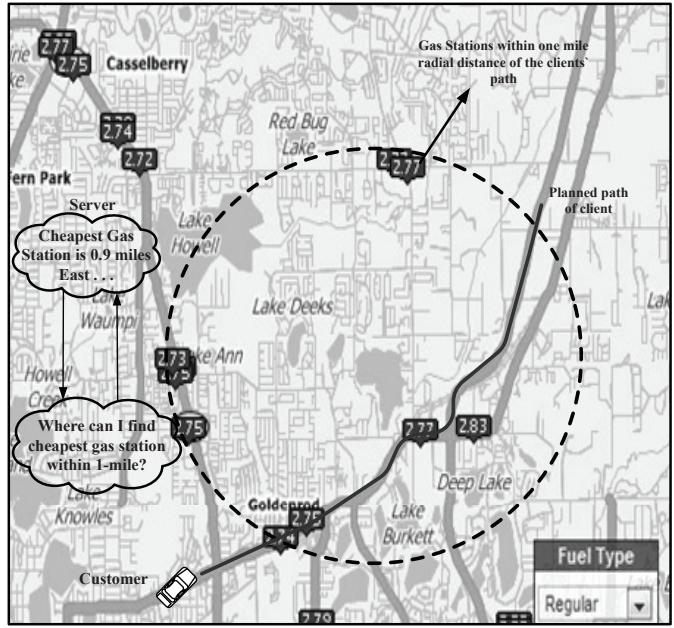} } 
\caption{Example scenario of interaction between a client and the GPIT} 
\label{fig:Example} 
\end{figure}

To illustrate the operation of the system, let us consider the example
in Figure \ref{fig:Example}. A driver travels from work to home along
the marked path. Near Lake Burkett, the car signals low fuel. The driver
has several nearby choices for refueling: some of them are on his
planned path, while others require short detours. He contacts the GPIT and requires information about the cheapest option in the vicinity.
The GPIT obtains this information from its local database
which contains information it has acquired ahead of time from drivers.
It will provide this information for a price, which will be negotiated in real time. 

Let us summarize the interests of the parties in this economic model:

\begin{enumerate}[i)] 

\item The revenue of the GPIT is earned by selling
information about the most advantageous gas buying location near a
trajectory. It incurs the cost of buying the information necessary
to keep the database up to date. The GPIT tries to maximize its profits
over the long term.

\item The drivers, in the role of clients, are acquiring information to
save on the cost of gas. The amount of savings depends on many
parameters: the variation of the price among gas stations, the detour
necessary to reach the cheaper gas, the size of the gas tank, and so on.
The savings are always relative to the drivers default preference: if
the GPIT instructs the driver to go to the gas station where it would
have gone anyhow, the savings are zero.

The cost of the information will be negotiated by the driver and the
GPIT. The challenge here is that the parties need to negotiate without
knowing the savings which will be provided by the information.

The fact that the information was worthless in one instance does not
guarantee that it will be the same the next time around as well. Our
approach will be to allow the client to estimate its savings considering a longer history, and to evaluate the benefits of dealing with the GPIT
over a longer timeframe as well.

\item Drivers selling price information to the GPIT are trying to
maximize their income. Their costs are the cost of making the
observation, communication costs and the cost of privacy (as they need
to disclose their location and verify their identity).

\end{enumerate}

From the point of view of the cost structure of its business, the GPIT
is similar to a software company: it has an initial cost to acquire the
information, but subsequently, it can sell it an arbitrary number of
times, without incurring further costs (as long as the database is
accurate).

What makes the GPIT business challenging is the difficulty in
determining the utility of the information to the user, the very large
variation of the utility from user to user, and, as we shall see, the
variation of the utility over time periods as well.

These considerations require us to consider a system where the cost of
information is determined on a case by case basis, by a short
negotiation process between the user and the GPIT. Not all negotiations
will be successful. In fact, there will be clients who will effectively
drop out from the system. This will happen if the clients estimate the utility of the information provided by the GPIT to be zero (considering the driving habits and location of the client). The subset of clients who will regularly purchase information depends on the pricing and negotiation strategy of
the GPIT. As we shall see, specific long term pricing strategies such as
promotional offers are also an important part of the system.

To understand the behavior of this system, we need a {\em detailed
simulation}. The system is only functional if it provides long term
profits to all its participants. But the profits of participants can
depend on very arcane details: the driving habits of the user, the
variation of gas prices in a specific area, the default fueling habits,
the geographical distribution of the gas stations, the estimate of the
GPIT of the user's profits, and so on. We also need to consider the
social behavior of the user: will it even consider negotiating again if
for several consecutive times the information was useless?

This paper presents the description and experimental validation of such
a detailed simulation. We use real world geographic data,
lifestyle-dependent driving patterns and vehicle models to create an
agent-based model of the drivers. We use real world statistics of gas
price fluctuation to create scenarios of temporal and spatial
distribution of gas prices. The price of the information is determined
on a case-by-case basis through a simple negotiation model. The trader
and the customers are adapting their negotiation strategies based on
their historical profits.

We are interested in the general properties of the emerging information
market: the amount of realizable profit and its distribution between the
trader and customers, the business strategies necessary to keep the
market operational (such as promotional deals), the price elasticity of
demand and the impact of pricing strategies on the profit.

The remainder of this paper is organized as follows.
Section~\ref{sec:SourceOfProfits} discusses the source of profits in the
system. Section~\ref{sec:Negotiation} describes the negotiation
protocol and the strategies used by the agents. Section~\ref{sec:dataset} describes the modeling of the environment and the market. An experimental study is shown in Section~\ref{sec:experiment}. Section~\ref{sec:RelatedWork} discusses related work.

%% file: profitsDiscussion.tex
\section{The source of the profits}
\label{sec:SourceOfProfits}

As we discussed, the economic framework of the GPIT requires that all the participants profit. Furthermore, all the participants express their profits in financial term (not in goods or utility). This requires that the system has a constant inflow of money.

The money in our system comes from the lost profits of gas stations. More specifically, the gas stations will loose those profits which are due to the customers lack of information about the variation of prices in time which are not consistent with the spatial distribution and perceived quality. 

Note that the lack of information is not always the reason for a customer to choose more expensive gas. Sometimes we
have a major brand with a local discount station next to each other,
with different advertised prices visible for both. Both of them
have clients because some clients are willing to pay for better
services and/or for better perceived quality of gas. This kind of price
difference is not a source of income for our system.

Similarly, temporal price variations which do not produce a reversal in
the ranking of optimal fueling station are also not a source of profit.
If a specific gas station is always the cheapest, this information has
no financial worth for a recurring customer.

\medskip

Let us consider an example where a gas station realizes profits based on the lack of information of its customers. Many airports are 
surrounded by gas stations which charge a substantial premium over the
prices prevalent in the metropolitan area. These gas stations take
advantage of the car rental companies' requirements that the rental cars
be returned with a full tank.

The rental car customers are frequently not well acquainted with the area and have difficulty identifying the closest gas station with advantageous prices. This lack of information is an important component of the gas station's pricing strategy; there are examples \cite{price-gauging-orlando} of a gas station
suing the city for the right {\em not} to post its prices.

The major price differences for the airport gas stations are an extreme
case. Yet, statistical data shows that gas prices offer sufficient temporal and spatial variability to provide an income to the GPIT economic system.

It is unclear whether the changing shopping patterns in a zone where a
large majority of buyers use the GPIT system would eliminate the
arbitrage opportunity. There is a reason to assume that only a subset of
drivers would participate in such a system, which would keep the
information trading opportunities functional.

%% file: Negotiation.tex
\section{Negotiating the price of information}
\label{sec:Negotiation}

% ********************************************************
%
%   The negotiation protocol
%
% ********************************************************

\subsection{The negotiation protocol}

Let us consider a customer taking daily trips dictated by its schedule
and lifestyle, gradually exhausting the fuel in his car. When the
quantity of fuel drops below a certain level the customer will look for
a gas station to refuel. The current location and planned path of the
customer has a significant impact on which gas stations are the best
refuelling choice.

To save on the cost of fuel by choosing the optimal gas station, the
driver contacts the GPIT, sends its location, planned path, and requests
information about the optimal gas station at the same time making a
price offer $x^1_{c \rightarrow s}$. The GPIT calculates this
information and replies with a counter offer $x^2_{c \leftarrow s}$.

The negotiation $\mathcal{X}^{m}_{c \leftrightarrow s}$ will consist of
a series of exchanged offers $\{x^{1}_{c \rightarrow s} , x^{2}_{s
\leftarrow c}$ $\cdots x^{n}_{c \leftrightarrow s}\}$. If the
\textit{kth} negotiation was successful, it will conclude with a deal
$v^{k}$. In this case, the client will pay $v^{k}$ dollars to the GPIT,
and the GPIT will deliver the information to the client (who will use it
to choose the gas station to refuel). If the negotiation concludes with
the conflict deal, no money or information is exchanged and the client
will refuel at the first gas station on its projected path. A full
negotiation round without concession from either side breaks the
negotiation.

% ********************************************************
%
%   The customer's negotiation strategy
%
% ********************************************************

\subsection{The customer's negotiation strategy}

The negotiation strategy of the client is a monotonic, uniform
concession with reservation price. The first offer made by the client is
at $u_{min}$, it will uniformly concede with a concession rate $\delta
\in (0, 0.5]$ until it reaches the reservation price $u_{max}$, at which
point it will insist on its offer. 

The customer's negotiation strategy is framed by its expectations of
benefits. At the beginning of the negotiation the client will set a
reservation price $u_{max}$ which is the maximum limit it is willing to
pay for the information. A natural limit would be the benefit the user
could acquire at this moment: yet, not having an information about the
gas prices, the user can not exactly know this information.

The user estimates that the information provided by the GPIT will save,
in average, the same amount it has saved in the past. It is not,
however, a good idea to set the reservation price $u_{max}$ to the exact
estimate, because this will terminate some negotiation which, after the
real savings have been computed, would have turned out to be profitable.
With this strategy, the deal is $x^{n}_{c \leftrightarrow s} \in
(u_{min} , u_{max}]$.

The conflict offer set by the customer is twice the initial offer and is
normally distributed around the initial offer, i.e., $u^{c}_{max} = 2 .
u^{c}_{min} + \mathcal{N}(\mu, \sigma^{2})$ where where $\mu$ = 0 and
$\sigma^{2}$ = 1. Therefore, the cost of a deal is given as
\[
 cost(\mathcal{X}^{k}_{c \leftrightarrow s}) = v^{k} =
  \begin{cases}
   x^{n}_{c \leftrightarrow s} & \text{if   }x^{n}_{c \rightarrow s} \geq u^{s}_{min} \lor x^{n}_{c \leftarrow s} \leq u^{c}_{max} \\
   0 & \text{otherwise}
  \end{cases}
\]

At the start of each negotiation the initial offer is dependent upon the previous $r$ negotiations with the GPIT:
\begin{equation} 
u^{c}_{min, k} = [m \cdot \frac{\sum_{n=1}^{r} v^{k-n}}{r}] + u^{c}_{min, k-1} \cdot (1 - m) 
\end{equation} 

\noindent where $m$ = 0.125 bounds the initial offer to increment within
one fourth of previous negotiated deal. The limitation of the amount of
history is justified by the experimental fact that while recent
negotiation experience can be a good predictor of the current
negotiation, older negotiations are not.

A parameter $c_{max}^{u}$ limits the size of $u^{c}_{min}$ such that the
agent does not start with a too high initial offer.

% ********************************************************
%
%   The GPIT negotiation strategy
%
% ********************************************************

\subsection{The trader's negotiation strategy}

The GPIT also uses a reservation price-based monotonic concession model as its negotiation strategy. Its reservation price and concession step is determined for each client and each negotiation using an adaptive learning process which considers the history of previous negotiations. 
It uses exponential smoothing to give more weight to the
recent deals. Exponential moving average helps smooth out
the previous deals and provides GPIT with a trend following indicator
for the future negotiations. The initial offer from GPIT in start of
$kth$ negotiation is
\begin{equation}
u^{s, k}_{max} = m \cdot \phi^{k}(r) + u^{s, k-1}_{max}(1 - m)\
\label{eq:GPITOffer}
\end{equation}

\noindent where m = 0.125 and $\phi^{k}(r)$ is the exponential moving
average of previous $r$ deals:
\begin{eqnarray}  
\phi^{k}(r) = \alpha \cdot v^{k-1} + (1-\alpha) \phi^{k-1}(r)
\label{eq:expAvg}
\end{eqnarray} 

\noindent where, there control factor is chosen as $\alpha$ = 2 / $(r +
1)$ providing more weight to the recent negotiation.

\medskip

If a customer is satisfied with the savings provided by the deals in
preceding negotiations, it will concede more in following negotiations.
To model the gradual change in the customers' concession, the GPIT uses
\textit{Moving Average Convergence Divergence} $\phi^{k}_{\varphi}(r)$
(MACD) \cite{appel1979moving}. This indicator provides a histogram with
the help of which GPIT estimates the concession behavior of the
customer. The histogram is obtained by computing the difference of two
EMAs (of different weights) against an EMA of the difference. The signal
$\varphi$, i.e., difference between two EMAs is given as:

\begin{equation} 
\varphi = \phi(9) \mbox{ }{-}\mbox{ }\phi(15)
\label{eq:EMADiff}
\end{equation}

The choice of using EMA's of period 9 and 15 provide us with two
different windows required to compute the recent shift in momentum of
negotiations. Evaluating $\varphi$ with brackets of 9 and 15 takes-in
consideration the current negotiations as well negotiations preceding it
and these values have been determined experimentally to provide a good
match of the customer behavior (in stock market analysis the values of
9, 12 and 26 day moving averages are used). The concession step of GPIT
$\delta^{c}_{k}$ is dependent upon $\varphi$ and is given as

\begin{equation} \label{eq:MACD}
\delta^{c}_{k} = \begin{cases} 0.15 & \mbox{if   } \varphi <
\phi^{k}_{\varphi}(5)
\\  0.25 & \mbox{else }\end{cases} 
\end{equation}

\begin{figure}[h!] \centerline{
\includegraphics[width=\columnwidth]{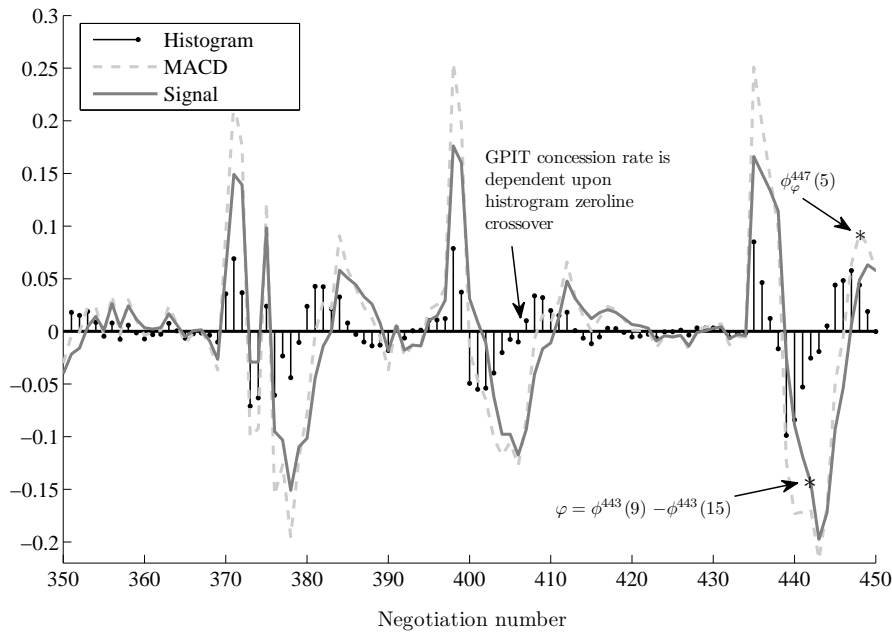} } 

\caption{Estimation of customer concession policy using MACD}
\label{fig:MACD} \end{figure}

In Figure \ref{fig:MACD}, the signal $\varphi$ shows the corresponding
divergence or convergence of the faster reactive $\phi(9)$ EMA from
slower reactive EMA $\phi(15)$. MACD histogram models the gradual shift
of the customer behavior and outputs a positive indication on histogram
if the customer is conceding more. The GPIT responds to the faster
concession rate of the customer by lowering its own concession rate
according to Equation \ref{eq:MACD} (and the reverse).

%% file: dataset.tex
\section{Environment modeling}
\label{sec:dataset}

As we have outlined, the economic model of GPIT is highly sensitive to
the environment conditions under which it operates. Profit margins are
thin, at best, the service might be useless for some clients, and
depending on certain, rather subtle conditions, the whole economic model
might be inoperable.

To account for these factors, the environmental conditions under which the system operates have been modeled based on real world data (and extrapolations of it) and at a relatively high level of detail.

\subsection{Modeling geography} 

We have chosen to model a relatively
large area of East Orlando. This area has a number of employers in
education and high tech industry, as well as residential, shopping
and recreation areas. It is a ``driving city'': public transportation
options are virtually non-existent, the distances between areas are
large and the zoning restrictions frequently prevent the building of
mixed use areas.

Our simulation uses the real world location of major employers in the
area, major residential and shopping areas. These areas have been
identified as locations on Google Maps. The exact locations of the
gas stations have been also entered in the database. Driving
directions and trip times have been acquired from the Google Maps. The simulator was directly contacting the web service, parsing the returned KML files, and caching them locally for reuse. 

\subsection{Modeling lifestyle}  

The lifestyle of East Orlando residents consists of well distinguished
time periods spent at distinct locations, and driving between them. We
have modeled the lifestyle of our simulated clients by randomly
assigning them workplaces and residences. The clients were assumed to
have an 8-to-5 work schedule on weekdays, not going to work on weekends,
and a randomly distributed set of other activities in the remaining
time. 

\subsection{Scenario generation}

The simulation of the economic system starts with a baseline scenario
containing the movement patterns of the clients, over the course of a
year, assuming that no refueling is needed. We assume that the
variations of the price of the fuel do not impact the in-city driving
patterns of the users.

Running the economic model on this scenario will add the refueling
times, occasionally inserting small detours necessary to find the
cheapest gas station.

The generated scenario is implemented as a series of events which
represent driving activities happening at specific locations and time.

Each event is described by (a) the participating agent, (b) the event
type, (c) the location, (d) the timestamp and (e) the distance travelled
since the previous event. An example subset of the scenario is shown in
Listing 1.

The following event types are supported:

\begin{itemize}

\item[-] ``departs'': the client departed a location

\item[-] ``arrives'': the client arrived at a location

\item[-] ``sees'': the client passes near a landmark (typically, for our
simulation, a gas station

\item[-] ``needs'': the vehicle needs fuel. This event is not in the
baseline scenario, but it is generated by the simulator

\end{itemize}

The location of the event is described by identifiers with standard
prefixes. These identifiers are then mapped to geographical locations by
a separate database. The prefix identifies the type of the location as
follows: W - work, S - Shopping, R - residential, STA - gas station.

The travel distance, used for ``arrives'' events, is calculated using
Google Maps queries for a suggested route from the latest departure
location to the current arrival location. The ``sees'' events are
generated by comparing the landmarks of STA type against the generated
path.

The simulator maintains the quantity of gas available for the cars
assuming a fuel tank of 15 gallons, an average consumption of 25
miles/gallon. The ``needs'' event is generated whenever the available
fuel dips below 2 gallons. The location of the ``needs'' event depends
on the negotiation process with the GPIT, the position of the refuelling
and the possible detours taken by the drivers to refuel. We assume that
the driver always completely refills the tank.

\subsection{Modeling gas prices}

\begin{figure} \centerline{
\includegraphics[width=\columnwidth]{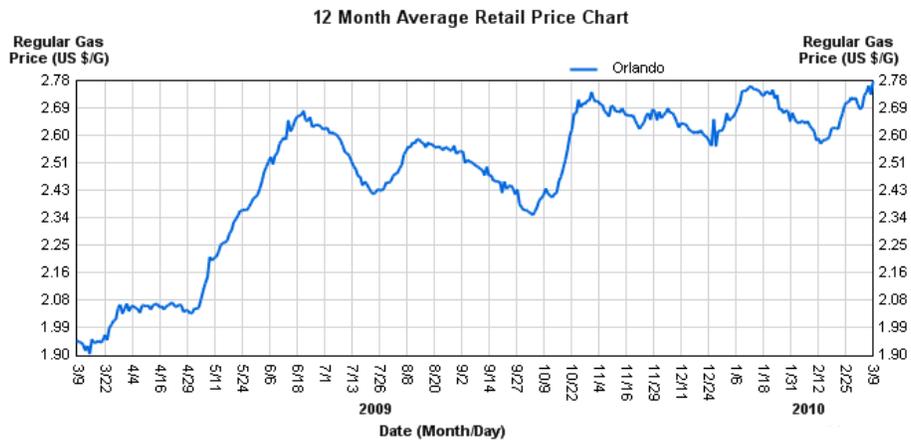} } 
\caption{Gas price trend for the year 2009-2010  (Source-GasBuddy)} 
\label{fig:gasChart} 
\end{figure}

We have acquired our gas prices from the website GasBuddy.com, which
works as an urban portal for people who can participate in posting and
reviewing gas prices. Based on these prices (see Figure~\ref{fig:gasChart}), we were able to generate a
number of fictional scenarios. To extrapolate the values we used a
technique to scale the spatial variance of the gas prices with an
arbitrary factor, and added a Gaussian noise component.

\begin{figure}
\centerline{
\includegraphics[width=\columnwidth]{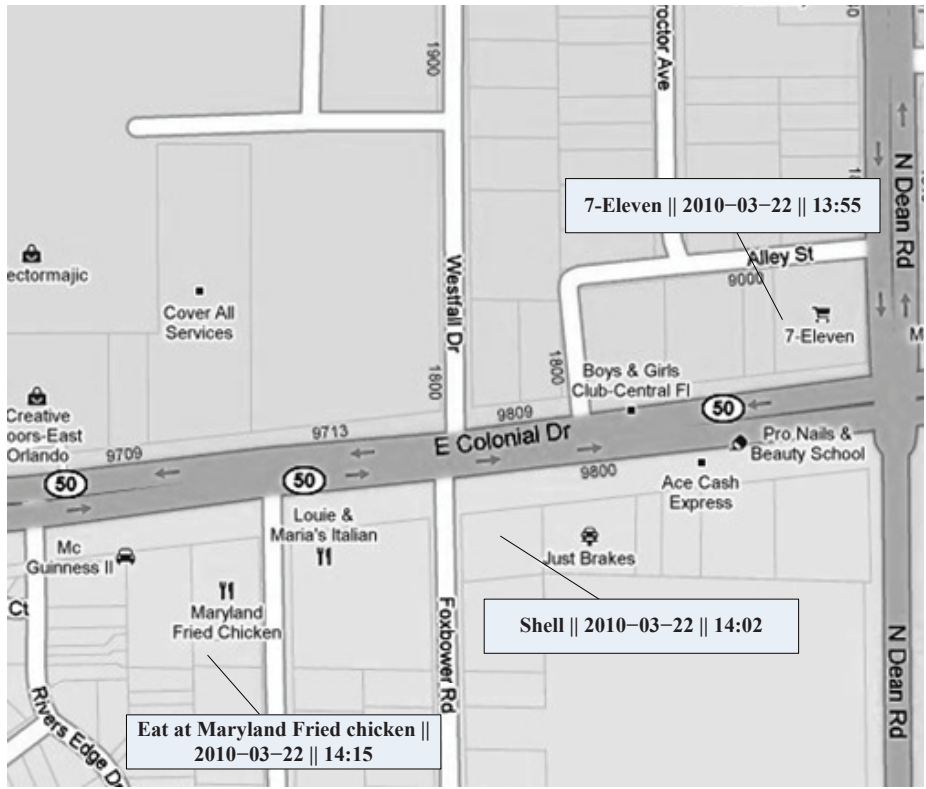} } 
\caption{Scenario Generated Map} 
\label{fig:Scenario} \end{figure}

%% file: experiment.tex
\section{Experimental study and results} 
\label{sec:experiment}

Our aim is to perform agent-based simulations to observe the general
properties of the information market. The GPIT simulator operates on the
assumption that every customer contacts the GPIT when gas refill is
required. For analyzing the properties related to information market and
GPIT we performed simulations, observing the following quantities:

\begin{itemize}

\item Total income for the server

\item The agreement price of negotiations

\item Number of successful negotiations

\end{itemize}

\subsection{Dynamic pricing analysis}

\begin{figure}[h!]
\centerline{
\includegraphics[width=\columnwidth]{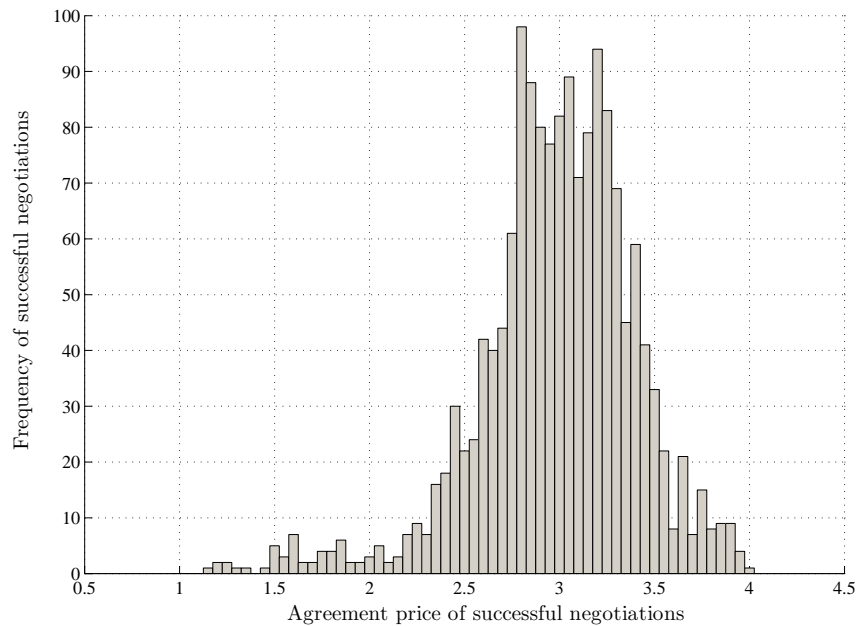} }
\caption{Dynamic pricing analysis} 
\label{fig:MACDFreq} \end{figure}

Our first series of experiments verify if the dynamic price formation mechanism operates as expected. Our expectation is that the strategies deployed by the customers and the trader interact in such a way that they will agree on a price level where both the customers and the trader will profit over the long run. Figure~\ref{fig:MACDFreq} shows the histogram of the distributions of the agreed price for successful negotiations. We find that the prices show a relatively wide distribution around the value of \$3 / information. This is consistent with the variation of the gas prices and the potential savings of the customers filling their 15 gallon gas tanks. 

\subsection{Price elasticity analysis using fixed initial price}

\if notinclude
The following settings were used for GPIT:
\begin{itemize}
\item Fixed initial Offer, $u_{max}^{s} = \$0 \mbox{ to } \$7$,
\item Conflict offer, $u_{min}^{c}$ = $u_{min}^{c}/3$, 
\item Concession rate, $\delta^{c}$ = 0.25
\end{itemize}
\fi

\begin{figure}[h!]
\centerline{
\includegraphics[width=\columnwidth]{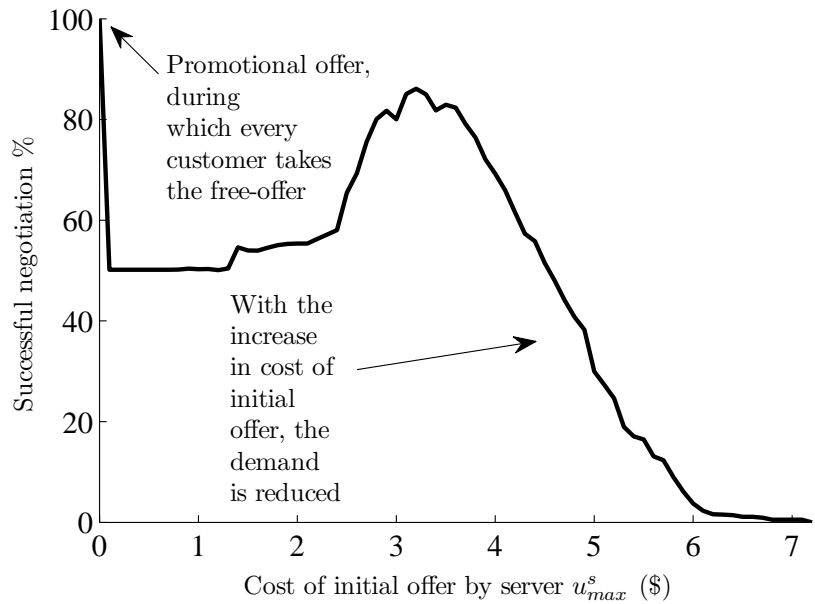} }
\caption{Price elasticity analysis} 
\label{fig:PercentSuccess} \end{figure}

Figure \ref{fig:PercentSuccess} shows the ratio of successful negotations for a fixed initial offer by the trader ranging in the interval $u_{max}^{s} = [\$0 .. \$7]$. The conflict offer was set to  $u_{min}^{c}$ = $u_{max}^{c}/3$, and the concession rate was fixed at  $\delta^{c}$ = 0.25. 

For the information offered for free (i.e. the promotional offer), the negotiation is always successful. Paradoxically, the highest success ratio is obtained at initial price settings which match the approximately \$3 value which is the average value of the dynamically determined price. 

\subsection{Profits obtained using fixed prices}

\begin{figure}[h!]
\centerline{
\includegraphics[width=3.75in]{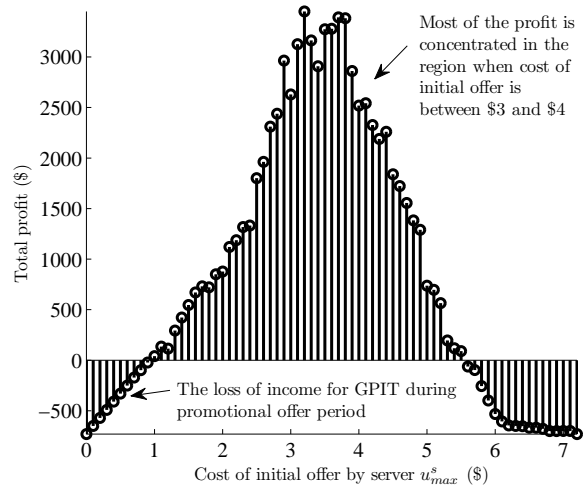} }
\caption{Total profit vs fixed initial offer cost} 
\label{fig:Profit} \end{figure}

Figure~\ref{fig:Profit} shows the profits of the trader obtained using a fixed initial offer. We observe the {\em rachet effect} \cite{acquisti-conditioning}  where even at higher prices some customers are willing
to pay for the price of information yet the total profit is very low. If
we compare the profits that are obtained using dynamic pricing as seen
in Figure \ref{fig:MACDFreq}, we see that it is not feasible for the
GPIT to operate using non-variable prices.

%% file: relatedWork.tex
\section{Related work}
\label{sec:RelatedWork}

\subsection{Agent-based markets} 

Kasbah \cite{chavez1996kasbah} developed
by MIT Labs, was one of the first and simplest agent based market in
which customers delegate agents on their behalf to do one-to-one
negotiations for market goods. In \cite{chavez1997real} the same group
used a slightly modified version of their previous market-place
implementation and performed real-life experiments. Sim
\cite{ren2009adaptive, sim2002market} discusses a model for negotiating
agents that is driven by dynamic markets. In his framework, he argues
that dynamic-market requires different tactics when compared to static
market and hence, variable adapting behavior should incorporated for
dynamic-market. The authors of Genoa \cite{raberto2001agent} provide an
agent based simulated financial market place. They assign certain
probabilities to a trader for buying or selling. The traders are merged
using graph theory, where any pair of randomly selected traders form a
cluster. The decision of agent in this market is constrained by
avaliable resources where the total avaliable cash is conserved.
NetBazaar \cite{sairamesh2009netbazaar} provides a distributed
marketplace for information products. Besides providing an
infrastructure for information market they have also provided the
support of federation. The main idea behind federation is to link
different trade systems for information sharing and perform trading on
behalf of each other. An agent based urban economic market is presented
in \cite{parker2008conceptual}. The design involves interaction between
multiple customers and sellers where the price is based on variance in
neighbourhood locality and other factors like spatial properties. An
agent based commodity trading scenario was demonstrated by authors in
\cite{cheng2009agent}. Though the authors were able to simulate the market scenario based on a single commodity yet it lacked the real world data. The
simulations were completely based on fictitious values by using
statistical methods.

\subsection{Price discrimination in information markets}

Varian \cite{varian2004economics} discusses different forms of price
discrimination that takes place in an information market. Initially he
discusses ``mass customization'' or ``personalization'' where price of
information is solely dependent upon a monopolist seller. Amazon was
accused of using similar variable pricing \cite{odlyzko2003privacy}. The
second type of discrimination also known as ``product line pricing'',
``versioning'' or ``market segmentation'' uses the distribution of consumers feedback for conditioning the price. The third discrimination is widely used and based upon selling information at different prices to different groups. We use the form of price discrimination that is based on conditioning using deal price history \cite{acquisti-conditioning}. 

\subsection{Promotional offers in information markets} 

Promotional offers holds significance in information marketplace
\cite{rowley1998promotion}. Promotional offers involve a considerable
amount of investment to gain attention of desired customers. This can
subsequently turn into profits if customers find the information
useful.

%\subsection{Agent based e-commerce market} 

%Researchers at Minnesota proposed an architecture MAGMA
%\cite{tsvetovatyy1997magma}, which provides an agent based market place
%for e-commerce.

%% file: conclusion.tex
\section{Conclusion}

This paper described the agent-based simulation of an information trading business where a trader agent buys and sells information about savings opportunities in gas prices. This is a low margin business where only a subset of customers can benefit from the savings opportunity, and only with a favorable rate of the information cost. Our our approach was to model the system in relatively high detail, including the geography of the area, the lifestyle choices of the customers, the spatial and temporal distribution of the gas prices. We have assumed that the agents are negotiating the price of the information on a case-by-case basis; their negotiation strategy is affected by their negotiation history. A promotional offer is used to establish the initial history necessary for efficient negotiation. 

Our experimental study shows that the proposed economic model is viable. The negotiation approaches, as designed, are efficient in finding the price ranges which ensure profits for both the trader and the clients, and the price/profit curve shows stable shapes with a maximum profit for the trader for a price of information around \$3.